\documentclass[10pt,a4paper]{scrartcl}   	
\usepackage{geometry}                		
\usepackage{graphicx}				
\usepackage{amssymb,amsmath}

\usepackage{mathptmx,helvet}
\usepackage[english]{babel}
\usepackage{booktabs}
\usepackage[version=3]{mhchem}
\usepackage[angew]{chemstyle}
\usepackage[numbers,super,sort&compress]{natbib}

\usepackage[backgroundcolor=green!40]{todonotes}

\usepackage{algorithmicx}
\usepackage{algpseudocode}
\usepackage{algorithm}

\usepackage[colorlinks, citecolor={black}, linkcolor={black}]{hyperref}

\usepackage{microtype}

\title{Modelling Chemical Reasoning to Predict Reactions}

\author{Marwin H.S. Segler, Mark P. Waller*\\
Institute of Organic Chemistry\\
and Center for Multiscale Theory and Computation\\
University Münster\\
marwin.segler@uni-muenster.de, m.waller@uni-muenster.de\\
Corrensstraße 40, 48149 Münster (Germany)
}

\begin{document}
\maketitle

\begin{abstract}
The ability to reason beyond established knowledge allows Organic Chemists to solve synthetic problems and to invent novel transformations. 
Here, we propose a model which mimics chemical reasoning and formalises reaction prediction as finding missing links in a knowledge graph. We have constructed a knowledge graph containing 14.4 million molecules and 8.2 million binary reactions, which represents the bulk of all chemical reactions ever published in the scientific literature. Our model outperforms a rule-based expert system in the reaction prediction task for 180,000 randomly selected binary reactions. We show that our data-driven model generalises even beyond known reaction types, and is thus capable of effectively (re-) discovering novel transformations (even including transition-metal catalysed reactions). Our model enables computers to infer hypotheses about reactivity and reactions by only considering the intrinsic local structure of the graph, and because each single reaction prediction is typically achieved in a sub-second time frame, our model can be used as a high-throughput generator of reaction hypotheses for reaction discovery.
\end{abstract}

Our innate ability to reason beyond established knowledge is one of the main driving forces of Science. \cite{klahr2002exploring,dunbar2005scientific,holyoak2012analogy,smith2012cognition} 
This ability allows Chemists to predict the outcome of reactions that have not yet been conducted. This is an essential skill in Organic Chemistry,  where synthesis and the search for new transformations are both fundamentally reaction prediction tasks.\cite{herges1990reaction} In reaction prediction, a prediction is made whether a reaction proceeds from reactants to the products in a forward manner.\cite{gasteigerbook}  It is of crucial importance to estimate the reactivity of the reactants in order to correctly predict reactions. However, reactivity can depend on the chosen  set of conditions and reagents,  and catalysts can even invert reactivity, as in Umpolung reactions.\cite{olga,guo2014n,campeau2008site,biju2011extending,bugaut2012organocatalytic} Therefore, reaction prediction more precisely involves predicting the products, catalysts, and reagents -- given only a set of reactants. 

Computer systems performing reaction and condition prediction are highly desirable, but are not yet used in mainstream organic chemistry.\cite{ley2015organic} For example, high-throughput computer-based reaction prediction (HTRP) can be used in de-novo drug design, virtual chemical space exploration or synthesizability estimation, or to predict whether disconnections proposed by a retrosynthesis system are feasible.\cite{chevillard2015scubidoo,warr2014short} HTRP with purely quantum-chemical or multi-scale approaches is yet to be conducted due to the tremendous computational cost of these methods. High-throughput reaction predictions are therefore performed with systems that use a model of chemical knowledge in some form or another. HTRP has been conducted with rule-based expert systems, machine-learning, formal logic and combinations thereof with force fields or semiempirical methods\cite{gasteigerbook,warr2014short,todd2005computer,kayala2011learning,kayala2012reactionpredictor,carrera2009machine,zhang2005structure,marcou2015expert,ramakrishnan2015big,ramakrishnan2015machine,ugi1993computer,forstmeyer1988reaction,herges1985synthesis,herges1992reaction,kowalczyk2009synthetic}: 

\begin{itemize}
\item Expert systems based on manually entered or algorithmically extracted reaction rules are the most widely used approach to predict reactions.\cite{warr2014short,todd2005computer,kayala2011learning,kayala2012reactionpredictor} They are appealing because they are interpretable and resemble the way undergraduate organic chemistry is taught. However, they suffer from several disadvantages: (a) They require knowledge to be manually encoded by experts either directly as reaction rules or indirectly as complex heuristics to extract rules from data, (b) become complicated to maintain with a growing knowledge base and (c) and they are not generalisable.\cite{kayala2011learning} 
Furthermore, they will only predict the chemistry encoded within the rules and thus cannot be used to discover novel chemistry.\cite{kayala2011learning,gothard2012rewiring2}%

\item HTRP with machine learning has recently regained traction with several interesting approaches.\cite{kayala2011learning,kayala2012reactionpredictor,carrera2009machine,zhang2005structure,marcou2015expert,ramakrishnan2015big,ramakrishnan2015machine} However, while these systems perform well within their intended domain of applicability, they are either limited in scope, or have not yet been shown to generalise to completely novel reaction types, especially in regards to transition-metal catalysis. In addition, they are not easily interpretable.

\item Ugi, Herges and co-workers pioneered the use of computers to invent novel reactions.\cite{ugi1993computer} They introduced the formal logic approach, which describes molecules and electron-shift in reactions as matrices, and applied it to discover novel pericyclic reactions.\cite{forstmeyer1988reaction,herges1985synthesis} Herges and Hoock invented novel [4+3]-pericyclic reactions by an exhaustive screening of formal reaction schemes.\cite{herges1992reaction} While the formal-logic approach is very well suited to invent pericyclic reactions, the authors stated that "transition-metal chemistry, however, is more difficult to describe with these methods."\cite{herges1992reaction} Apart from the formal logic approach, we are not aware of any high throughput computational approaches that have been proposed to discover novel reaction types.

\item Systems that predict reaction conditions are seldomly reported and are typically designed to predict specific reaction classes, for example the Michael-reaction.\cite{marcou2015expert}

\item  Baldi and coworkers reported an elegant approach to make predictions at the mechanistic level for reactions not involving transition metal catalysis.\cite{kayala2011learning,kayala2012reactionpredictor}. However, even though mechanisms undoubtedly play an important role in understanding chemical reactions, many reactions can be predicted just by their overall transformation. Detailed mechanisms for many reactions are not known, and reliable data is often unavailable. Furthermore, the reaction mechanism can even change based on the employed conditions.

\end{itemize}

Here, we introduce a novel approach for reaction prediction with the explicit aim of being able to generate high-quality hypotheses for novel transformations and reaction conditions. We hypothesised that to predict an unknown reaction, regardless of whether it is of an established or novel type, we have to find a missing link between reactant molecules in our current collective chemical knowledge. 
We will first give a formal description of our reaction prediction model, present the results of the validation studies, and then discuss the relationship between link prediction and chemical reasoning.

\subsection*{Molecule-Reaction-Graphs as a Knowledge Representation}
Knowledge about molecules and their reactions can be naturally represented as a graph. One could use a simple graph where molecules (nodes) are linked by reactions (edges). However, to allow for a more fine-grained representation of reaction roles, is it advantageous to use a directed bipartite graph $G=(M,R,E)$, which consists of molecule nodes $m_i \in M $ and reaction nodes $r_j \in R$ (see \ref{fig:chemtographfull} a,b)\cite{benko2003graph,ugi1979neue,grzybowski2009wired}
We extend this traditional representation by allowing edges $e_k = (m_i,r_j, t) \in E$ to have a type $t$, which represents the role $t \in \{$reactant, reagent, catalyst, solvent, product$\}$ of a molecule $m_i$ in a reaction $r_j$. If the edge has one of the first four possible roles, it is directed from a molecule towards a reaction. If the molecule is a product of a reaction, the edge points from the reaction to that molecule. 

\subsection*{Reaction Prediction as Link Prediction}

Reaction prediction can be generally defined as the task of predicting missing nodes and edges in a graph $G$ representing our current knowledge. To predict a reaction $r_x$ from a set of reactants $S$, the following steps have to be carried out:

\begin{enumerate}

\item Predict a node $r_x$, such that $r_x$ is linked to the reactants $m_i \in S$ with edges $e_i = (m_i, r_x,t)$ of the type \textit{reactant}. 

\item Predict  a node $m_p$  that is the product of the reaction with an edge $e_p = (m_p, r_x,t)$ of the type $t$ = \textit{product}. 

\item The necessary set $C$ of reagents, catalysts and solvents can be predicted by finding nodes $m_c \in C$ and edges $e_c = (m_c, r_x,t)$ of the type $t \in$ \{\textit{reagent,catalyst,solvent}\}. $C$ can be empty.
\end{enumerate}

To make a prediction about these missing nodes and edges, we can take inspiration from studies of link prediction in social and biological networks.\cite{liben2007link} One key finding in this context is that missing links in a network can be predicted by analysing the paths.\cite{liben2007link}  A path in a graph is a sequence of nodes $\pi^L_i = (n_0,n_1,...,n_L) $ of length $L$, in which each two subsequent nodes $n_i$ and $n_{i+1}$ are connected by an edge. Therefore, the first step we take to predict a reaction is to find paths between two molecules in the graph (in this paper, we will consider only paths along edges of the type \textit{reactant}). 
The length $L$ of a path $\pi^L=(m_1,...,m_n)$ can be used to indicate if molecules $m_1$ and $m_n$ are analogous or complementary (see Supporting Info Figure S2 for a scheme):
\begin{gather}
\pi^L(m_1,...,m_n) \Rightarrow \begin{cases} \text{complementary } &\mbox{if } L = 4n+2 \text{ where } n \in \mathbb{N}_0\\ 
\text{analogous } &\mbox{if } L = 4n \end{cases}
\end{gather}
Here we define molecules that (given specific conditions) can react with each other as \textit{complementary} in their reactivity. Analogous reactivity of two compounds is defined as the possibility to react with the same reaction partner(s). If no paths between two molecules can be found, we simply cannot make any statement about their mutual reactivity.

\paragraph{}
To ensure that retrieved paths are chemically meaningful, we define two filters:

\paragraph{Filter 1} We need to ensure that subsequent reactions in a path occur at common atoms. Therefore, a path $\pi^L_i = (...,m_{i-1},r_Q,m_i,r_{Q+1},m_{i+1},...)$ is only valid, if for every molecule $m_i$ in the path the atoms of $m_i$ that get changed in the preceding reaction $r_Q$ in that path and the subsequent reaction $r_{Q+1}$ are the same.  If we for example $m_i$ had two different functional groups, it would not be meaningful to compare the reactivity of $m_{i-1}$ and $m_{i+1}$ towards $m_i$, if reactions $r_Q$ and $r_{Q+1}$ occurred at different functional groups of $m_i$. 

\paragraph{Filter 2} We also need ensure that subsequent reaction nodes $r_J$ and $r_{J+1}$ in a path have similar reaction centres. The reaction center is the set of all bonds which are formed, changed or broken in the course of a reaction, and the atoms they connect. For this purpose, we employ reaction fingerprints because they have been shown to encode the reaction centre and to correspond to reaction classes.\cite{schneider2015development,de2012mining,carrera2009machine,latino2006genome,zhang2005structure} 
Fingerprints are vectors that encode if or how often certain features, for example a functional groups, are present in an entity.
The reaction fingerprint $\mathcal{F}$ of a reaction $R_i$ is a vector which is obtained by subtracting the sum of all $n$ counted fingerprints $p_j$ of the products and the sum of all $m$ counted fingerprints $r_k$ of the reactants:\cite{schneider2015development} 
\begin{gather}
\mathcal{F}(R_i) = \sum_j^n p_j - \sum_k^m r_k
\end{gather}
In this work, reaction fingerprints based on Extended Connectivity Fingerprints (ECFP4 and FCFP4)\cite{rogers2010extended}, as described by Schneider et al.\cite{schneider2015development} were employed, which were used as implemented in the Chemistry Development Kit (CDK) version 1.5.8.\cite{steinbeck2003chemistry} The fingerprints are then compared by calculating their \textsc{Tanimoto}\cite{gasteigerbook} score $T(A,B)$ for continuous values:
\begin{gather}
T(A,B) = \frac{\sum^{n}_{i=1} x_{iA} {x_{iB} }}{\sum^{n}_{i=1}  ( x_{iA}^2 +  x_{iB}^2 - x_{iA} {x_{iB}})}
\end{gather}

\paragraph{Chemical Reasoning Algorithm} A bi-directional path search is performed as a breadth-first search starting from two query molecules $M_i$ and $M_j$, see Algorithm 1. When the two branches meet, the resulting path is saved. During the path search, the algorithm calculates the reaction fingerprints of subsequent reaction nodes and their \textsc{Tanimoto}\cite{gasteigerbook} similarity. If the similarity is below a certain threshold, the path is excluded from the search. This threshold is the only parameter of the model, and was set to a value of 0.2 after a parameter search was carried out on a developmental subset of the data.  The algorithm continues to find more paths until there are no more possibilities to explore the graph further. The number of returned paths was limited to 1000 for performance reasons.

\begin{algorithm}[h]
\caption{\texttt{ Chemical Reasoning Algorithm}}
\label{alg:mentor}
\begin{algorithmic}[1]
   \State {\bfseries Parameters:} Tanimoto threshold $t$
  \State {\bfseries Input:} molecules $M_n$
    \If{the two branches meet}
 return the path from $M_i$ to $M_j$\Else

    \For{all reactions $R_n$ where $M_n$ is a reactant}
    
    \State retrieve $R_{n-1}$ 
  \Comment{preceding reaction of $R_n$ in branch}
  \If {\texttt{HaveCommonAtoms}($M_n, R_n, R_{n-1}$) }  \Comment{Filter 1}
  \If{ \texttt{Tanimoto}[\textbf{rfp}$(R_n)$, \textbf{rfp}$(R_{n-1})$] > $t$} \Comment{Filter 2}

\State{retrieve $M_{n+1}$ 
 \Comment{the reaction partner of $M_n$ in $R_n$}}
\State{call \texttt{Chemical Reasoning}($M_{n+1}$)}

  \EndIf
  \EndIf

   \EndFor
      \EndIf

\end{algorithmic}
\end{algorithm}

\paragraph{Product Prediction} To predict the structure of the products formed in a predicted reaction of two molecules $m_1$ and $m_L$, we use the concept of half reactions as described by Stadler and colleagues.\cite{benko2003graph,andersen2014generic}
Here, a binary reaction gets split into two half reactions, one for each reactant molecule, which also contain information about how the atoms and bonds of the reactant will be transformed and merged in the product. By combining two matching half reactions, we can get the "full" reaction.

If a valid path $\pi^L_i = (m_1,r_A,...,r_Z,m_L)$ has been found between two molecules $m_1$ and $m_L$ with a length L = 4n + 2 (complementary reactivity), then the structure of the possible product can be generated by obtaining the half reactions $h_1$ of molecule $m_1$ in reaction $r_A$ and $h_L$ of $m_L$ in reaction $r_Z$, respectively and merging $h_1$ and $h_L$ (see \ref{fig:chemtographfull} for a graphical explanation).

\paragraph{Condition Prediction} 

To predict reaction conditions, we again use the retrieved paths from the knowledge graph as follows: Let $\pi^L_i = (m_1,r_A,...,r_Z,m_L)$ be a path between two molecules $m_1$ and $m_L$ with a length L = 4n + 2. We then denote the sets of reagents, catalysts and solvents involved in the first reaction $r_A$ and the last reaction $r_Z$ in path as $C_A$ and $C_Z$ respectively. We then propose an initial set of conditions $C_p$ for the predicted reaction to be 
\begin{gather}
C_p = C_A \cup C_Z \label{condpred}
\end{gather}

\subsection*{Illustrative Examples}
\begin{figure}[h!]
\begin{center}
\includegraphics[width=\textwidth]{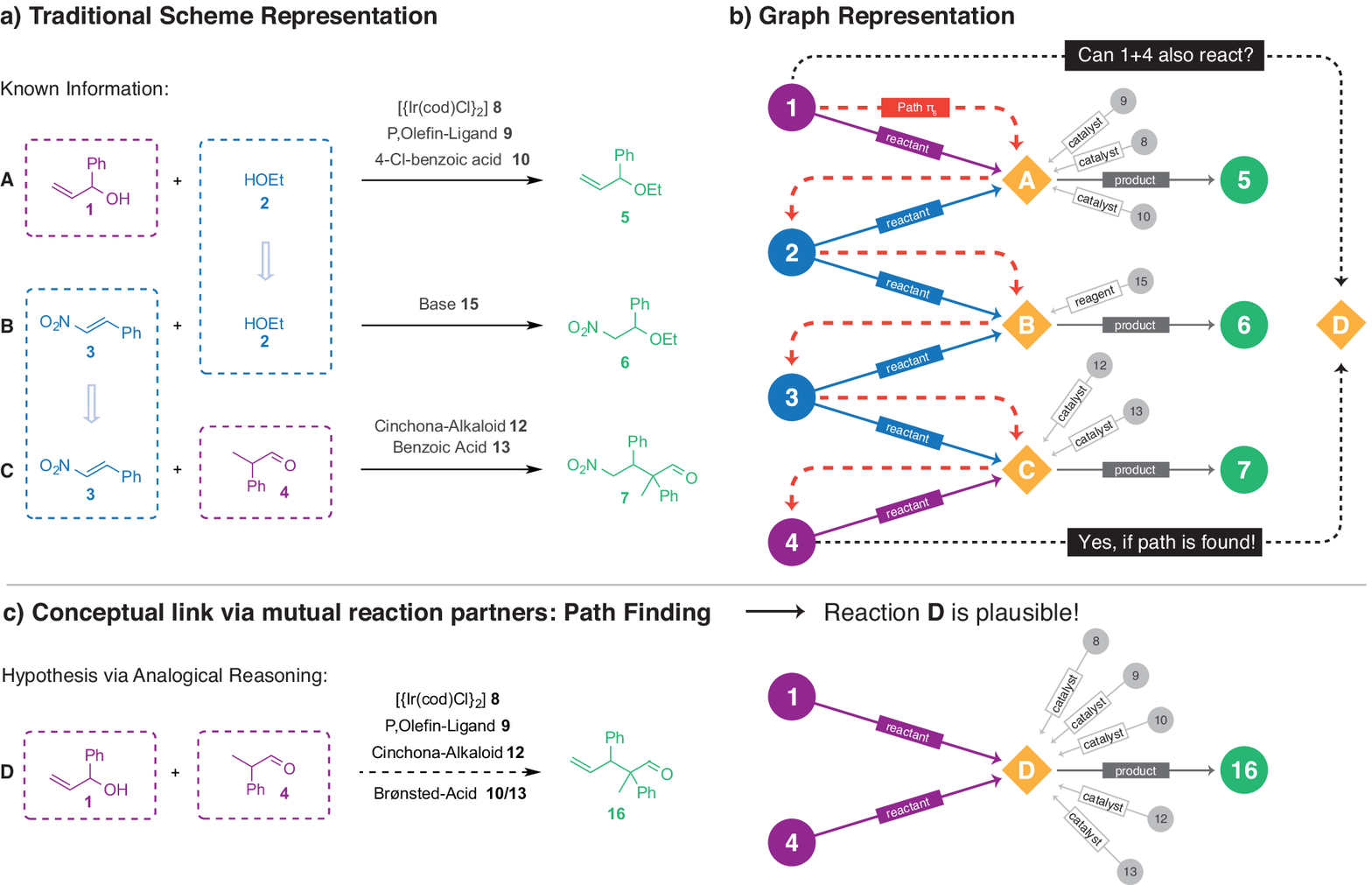}
\caption{A Scheme Representation a) can be translated into a b) Graph Representation. Numbered nodes correspond to molecules (circles), reaction nodes (diamonds) are designated with a letter. The graph representation allows a fine-grained encoding of the roles molecules can play in a reaction and can be used to study the relationships of molecules.
c) To perform reaction prediction, path search in the graph performed. To predict the reaction of \textbf{1} and \textbf{4}, we retrieve the path \textbf{1}$\rightarrow$A$\rightarrow$\textbf{2}$\rightarrow$B$\rightarrow$\textbf{3}$\rightarrow$C$\rightarrow$\textbf{4} (red dotted line). This corresponds to a logical explanation: Because \textbf{1} and \textbf{3} react with \textbf{2}, they have analogous reactivity, and \textbf{3} and \textbf{4} have complementary reactivity, \textbf{1} and \textbf{4} likely also react. From the retrieved path, we can predict the possible product and the needed catalysts and reagents by concatenation of half reactions (see left side).}
\label{fig:chemtographfull}
\end{center}
\end{figure}

To ensure the concepts in our model are clearly understood, we present a side-by-side comparison showing chemists notation on the left, to the graph representation on the right, see \ref{fig:chemtographfull}. Consider that two molecules \textbf{1} and \textbf{3} both react with another molecule \textbf{2} at the same atoms of \textbf{2}. From a chemical perspective, \textbf{1} and \textbf{3} thus have  \textit{analogous} reactivity. In the graph, this would correspond to the path $\pi^4_a = (m_1,r_A,m_2,r_B,m_3)$. Now, imagine \textbf{1} and \textbf{3} have many different mutual reaction partners $m_2$, so there would be a set $\{\pi^4_{z}, \pi^4_{y},...\}$ of different paths $\pi^4_{i}= (m_1,...,m_3)$. This would correspond to many different explanations why \textbf{1} and \textbf{3} are analogous in their reactivity and thus be an even stronger indicator than a single explanation.
Suppose we combine path $\pi^4_a = (m_1,...,m_3)$ with the knowledge that \textbf{3} reacts with another molecule \textbf{4} ($\pi^2_b = (m_3,r_C,m_4)$). Because it was established earlier that \textbf{3} has analogous reactivity to \textbf{1}, we can infer that \textbf{1} and \textbf{4} are likely complementary.  This corresponds to a path $\pi^6_c = (m_1,r_A,m_2,r_B,m_3,r_C,m_4)$ between \textbf{1} and \textbf{4} in the graph.  As above, if many paths with $L=6$ could be retrieved between the two molecules, this would give many different explanations for the feasibility of the possible reaction. If several paths between two molecules can be found, they can correspond to different reactions, but also to the same reaction performed under different conditions, which is reasonable, because two molecules can yield different products under different conditions.\cite{guo2014n,campeau2008site}

\begin{figure}[bt]
\begin{center}
\includegraphics[width=0.7\textwidth]{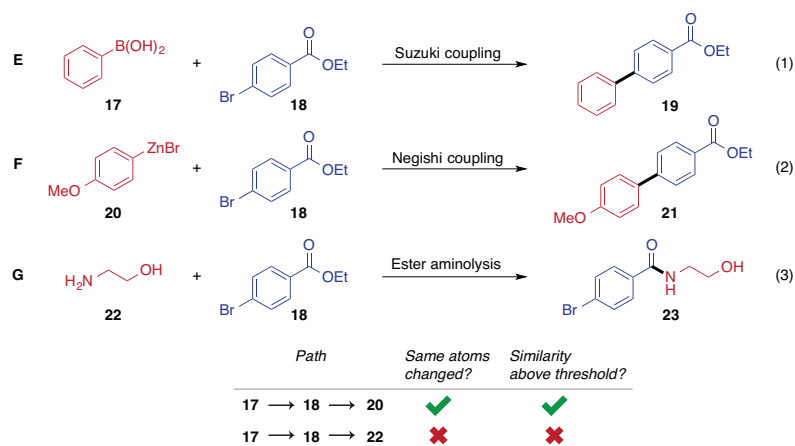}
\caption{Explanation of the two filters used: 1) Boronic acid \textbf{17} and the aryl zinc bromide \textbf{20}  both react at the bromide moiety of \textbf{18}. The same atoms of \textbf{18} get changed in reaction \textbf{E} and \textbf{F}. Additionally, the Tanimoto similarity of the reaction fingerprints of \textbf{E} and \textbf{F} is 0.41, which is above the threshold of 0.2. 
The path 
\textbf{17}$\rightarrow$
\textbf{E}$\rightarrow$
\textbf{18}$\rightarrow$
\textbf{F}$\rightarrow$
\textbf{20} can thus be used to infer the analogous reactivity of \textbf{17} and \textbf{20}.
2) In contrast, Aminoethanol \textbf{22} reacts at the ester moiety of \textbf{18}. \textbf{17} and \textbf{22} thus cannot be compared in their reactivity towards \textbf{18}. Also, the Tanimoto similarity of the reaction fingerprints of \textbf{E} and \textbf{G} is 0.0, which is below the threshold of 0.2. Thus, the path 
\textbf{17}$\rightarrow$
\textbf{E}$\rightarrow$
\textbf{18}$\rightarrow$
\textbf{G}$\rightarrow$
\textbf{22} is filtered out.
}
\label{fig:filters}
\end{center}
\end{figure}

Examples of the two chemical filters are shown in \ref{fig:filters}.

\section*{Results and Discussion}

\subsection*{Data}
We constructed a knowledge graph using 14.4 million molecules and 8.2 million binary reactions from Reaxys\cite{reaxys} published from the beginning of chemistry as a discipline until 2013. The dataset contains essentially all binary chemical reactions published in this period. It therefore features the complete spectrum of organic chemistry.
The reactions in the database are encoded in a human-readable way, and are sometimes not balanced. Since we were interested in building a model that handles real-world data, we did not perform any reaction balancing or cleaning. Chemical reactions were standardised by removing explicit hydrogens and mapped using Chemaxon Standardizer 6.2.1.

\subsection*{Quantitative Validation of Reaction Prediction}
 We performed a time-split validation, in which data gathered up to a certain point is used to predict data published after this point. It has been shown to give a more realistic estimation of classification performance in comparison to cross validation with randomized splitting and is closer to the desired objective of predicting future development.\cite{sheridan2013time} We used all  reactions from the Reaxys database that were first published in 2014 or later as our hold-out validation set. We then tested our model by using the reactants of a reaction as an input, and consider a reaction correctly predicted if one of the proposed products matches the reported product.

\begin{figure}[tb]
\begin{center}
\includegraphics[scale=0.5]{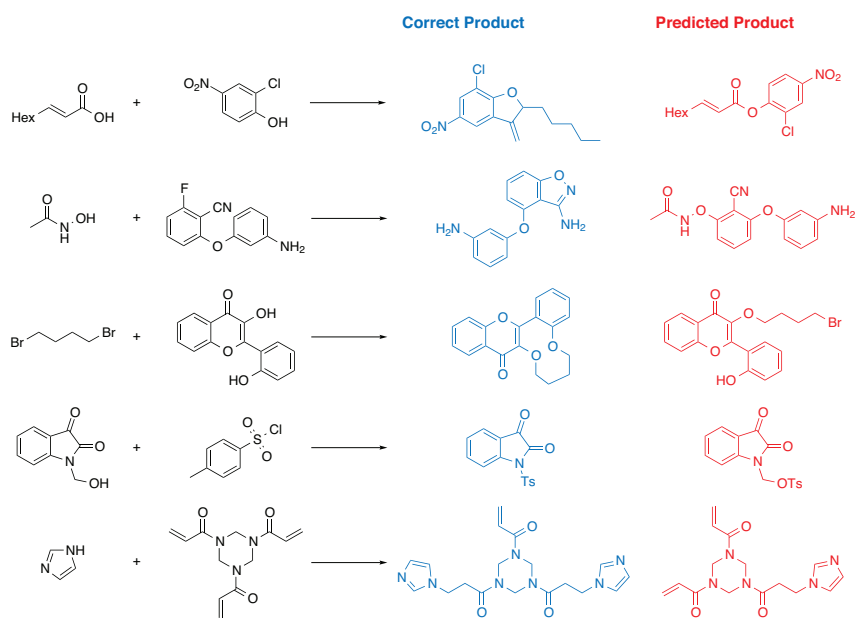}
\caption{Examples of falsely predicted reactions. However, the predictions are still chemically reasonable.}
\label{fig:failed}
\end{center}
\end{figure}

The results of our validation study are shown in \ref{tab:val}. Our algorithm correctly predicts the right products with a 67.5\% accuracy for 180,000 randomly selected reactions from our hold-out validation set. The median number of predicted products per reaction is 3, the mean is 5.3. Simply generating the products by combining all known half reactions of the two reactants without path search gives an accuracy of 78\%, which also represents the upper bound for our model. However, in this way a median of 62 and a mean of 311 products are generated for each reaction. Our path-finding based model thus efficiently limits combinatorial explosion and only proposes a few solutions for each problem. Our model does fail to predict reactions of a molecule in positions where it has not been activated before, or where a mechanistic consideration is necessary to predict the outcome. However, the failed predictions are still chemically reasonable. \ref{fig:failed} shows some examples, where our model fails to predict the correct product. 
In comparison, a transformation rule-based expert system as proposed by Christ et al.\cite{christ2012mining} and Law et al.\cite{law2009route} resulted in a 52.7\% accuracy for this validation set. This was achieved using 147755 reaction rules extracted from the reactions published until 2013.  To put these two accuracies into perspective, we show a baseline method (Entry 3), in which a random product is chosen from the list of the validation reactions. In this baseline method, $<$ 1\% of the reactions are correctly predicted. This highlights the extraordinary difficulty of the reaction prediction problem.

\begin{figure}[tb]
\begin{center}
\includegraphics[width=\textwidth]{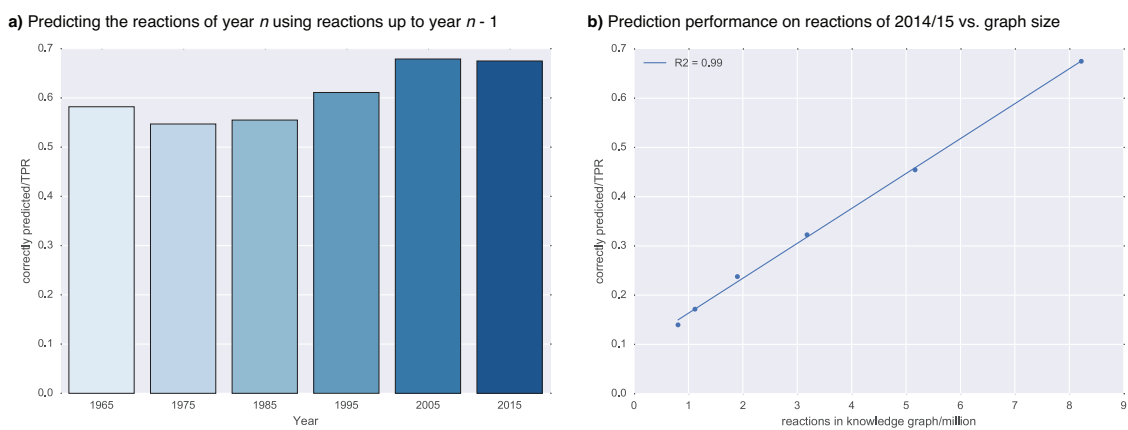}
\caption{a) Prediction performance on reactions published in year $n$, using reactions published until $n-1$ as the knowledge graph. The performance is consistent over the different years.
b) Performance on reactions published in year 2014/2015, using knowledge graphs of different sizes. Prediction accuracy increases linearly with the size of the knowledge graph.}
\label{fig:rxn-vs-acc-graph}
\end{center}
\end{figure}

To evaluate the performance of the algorithm at different points in time, we performed a series of time split-validation studies, predicting the reactions published in a certain year $n$ using all reactions published prior to that year as our knowledge base. \ref{fig:rxn-vs-acc-graph} a) shows that the performance is consistent over these different splits of time. Furthermore, we studied the dependence of prediction accuracy on the graph size (\ref{fig:rxn-vs-acc-graph} b). We found that the prediction accuracy on the validation set increases linearly with the knowledge graph size.

\begin{table}[htb]
\caption{Validation Results}
\begin{center}
\begin{tabular}{llr}
\toprule
Entry	 &Model	&Correctly Predicted\\
\midrule
1	&Our Model	& 67.5\% \\ 
2	&Rule based-Expert System	& 52.7\% \\
3	&Random Baseline &	$<$ 1\%  \\
\bottomrule
\end{tabular}
\end{center}
\label{tab:val}
\end{table}%

\subsection*{Predicting Novel Reaction Types}
Our model was derived with the intent to not just predict established and "common" reactions, but also to predict unprecedented reaction types. 
To test this ability, reactions that could be explained with transformation rules extracted from all reactions published until 2013 were removed from our validation set.

\begin{figure}[h!]
\begin{center}
\includegraphics[scale=0.6]{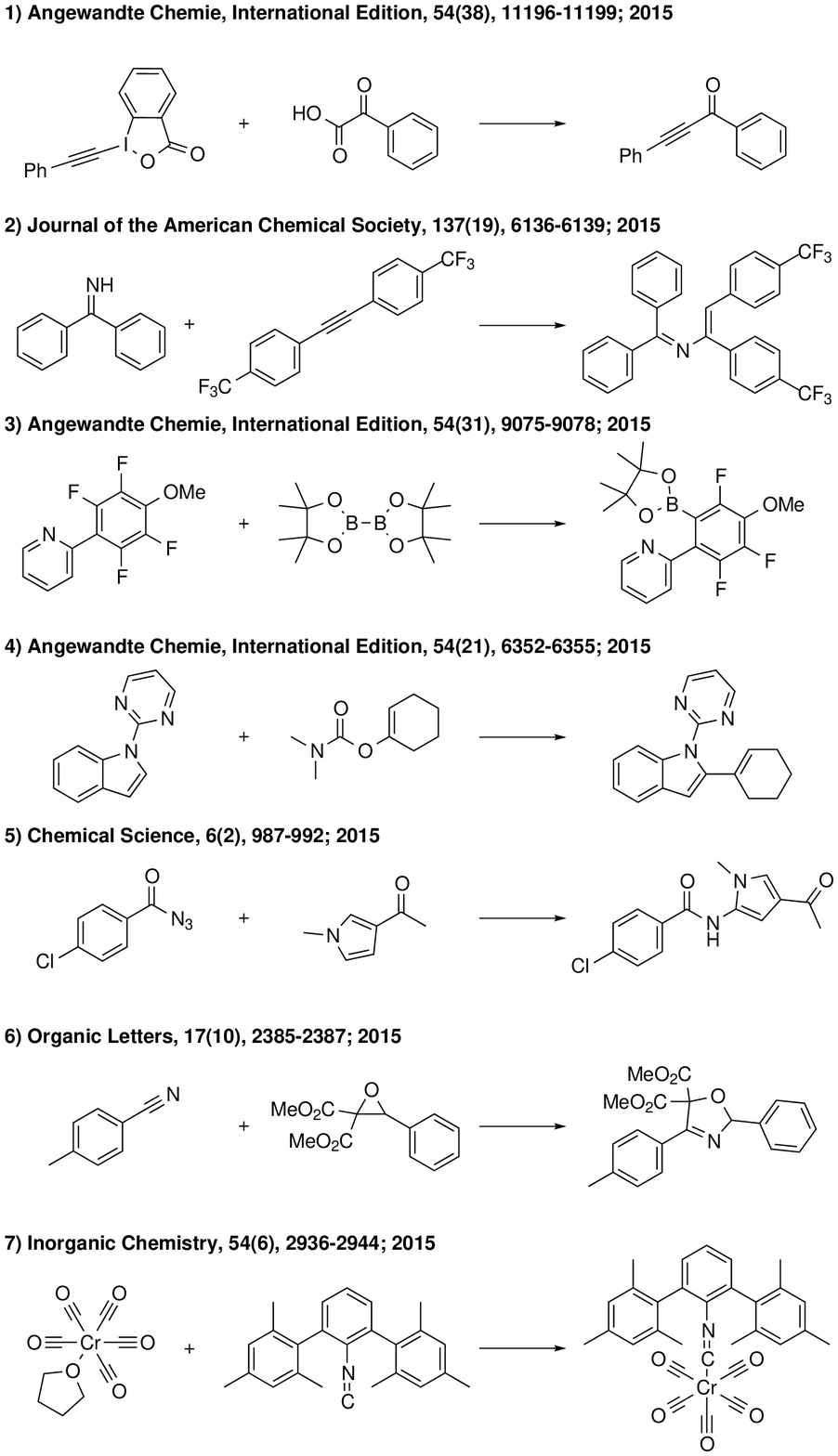}
\caption{Representative examples of successfully predicted novel reactions, with the respective references.
In this task the products were predicted given the reactants. Reagents and catalysts are omitted, since they were not predicted here.}
\label{fig:unprecedented}
\end{center}
\end{figure}

From the remaining reactions, a set of 13000 reactions was randomly selected.
Our algorithm predicts the correct products for 35\% of these challenging novel reactions. 
\ref{fig:unprecedented} shows some examples of unprecedented reactions that our algorithm could discover. Among them, there are for example state-of-the-art transition metal-catalysed C-H functionalisation reactions, which have been published in high-profile journals, but also other types of chemistry, and even organometallic reactions.

These results demonstrate that our model is capable of discovering new reactions. It also indicates that our graph-based approach complements rule-based systems, since these reactions cannot be predicted with existing rules by definition.

\subsection*{Negative validation}
While the positive evaluation of a reaction prediction system can be easily done with a test set of hold-out known reactions (as above), negative evaluation with reactions that are known \textit{not} to occur is a difficult task, because current publishing customs in organic chemistry do not provide incentives for publishing failed reactions or the limitations of synthetic methodology. This lack of data has been criticised both by synthetic chemists and chemoinformaticians.\cite{carrera2009machine,marcou2015expert,collins2013robustness}

To obtain data of reactions which are known \textit{not} to occur, we randomly selected 36000  known reactions from our validation set and generated "wrong" (but still plausible) products by the application of 94 hand coded reaction rules, covering the most common chemical transformations, to the reactants using rdkit.\cite{kurti2005strategic,hartenfeller2011collection} 
Our model was able to identify the wrong products and classify these reactions as not occurring in 94\% of the cases.
The model is thus able to prioritise the chemo- and regioselectivity of reactions very well. Again, this is achieved without explicitly encoded selectivity ranking rules, just from the provided reaction data in a knowledge graph representation. In contrast, in rule-based expert systems, functional group tolerance and competition between functional groups  first have to be encoded by hand or extracted with complex rules.

\subsection*{Empirical Validation of Condition Prediction}

To test the predicted reaction conditions and catalysts generated by employing equation (\ref{condpred}), we constructed a smaller knowledge graph containing 30000 reactions. This dataset was comprised mainly of catalytic reactions published in 2014 or earlier, and was annotated with reagents and catalysts. We chose 11 reactions from the recent literature as our validation set, and made sure that these 11 reactions were not contained in dataset of 30000 reactions. A quantitative analysis of condition prediction is difficult, because the chemical plausibility of the reaction conditions has to be judged by a chemist and cannot yet be automated. 
\begin{figure}[tb]
\begin{center}
\includegraphics[scale=0.5]{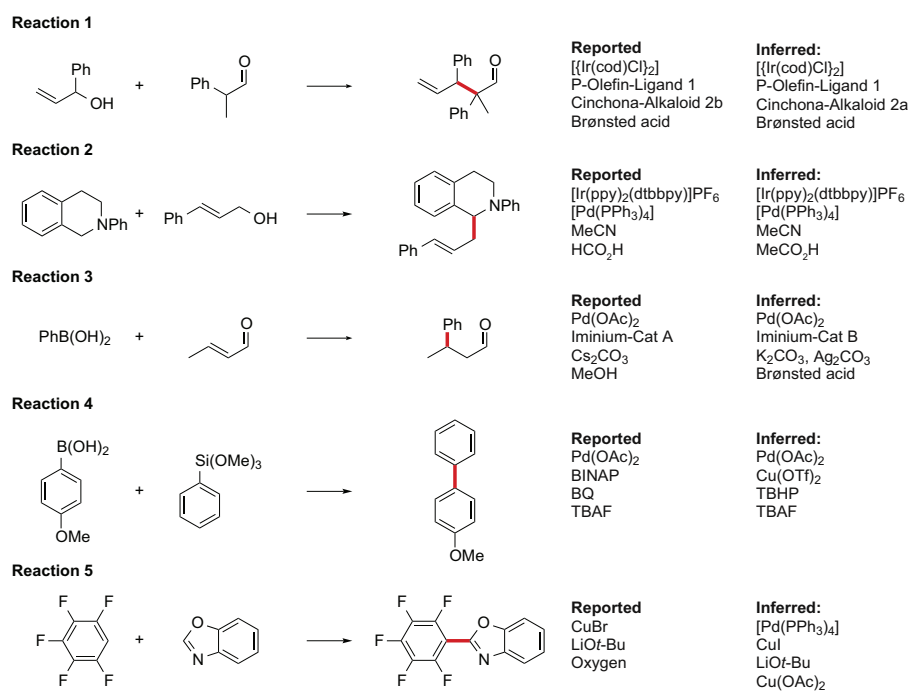}
\caption{Empirical Validation of Reaction Prediction. The algorithm generally proposes reasonable reagents and catalysts for the reactions. For details and more examples, see Supporting Information.}
\label{fig:condpred}
\end{center}
\end{figure}

\ref{fig:condpred} shows five examples, the complete results are listed in the Supporting Information. In all cases, the algorithm could predict the correct product. Encouragingly, the predicted conditions are similar to the reported conditions. In the cases where the conditions do not match, the algorithm proposes reagents and catalysts that are functionally similar, for example potassium carbonate instead of caesium carbonate as a base in reaction 3 or copper triflate and tert-butyl hydroperoxide (TBHP) instead of benzoquinone (BQ) as the oxidant in reaction 5. This shows that the algorithm is able to capture the chemical role of the involved reagents and catalysts remarkably well, again without any explicit encoding.

\begin{figure}[h!]
\begin{center}
\includegraphics[width=0.8\textwidth]{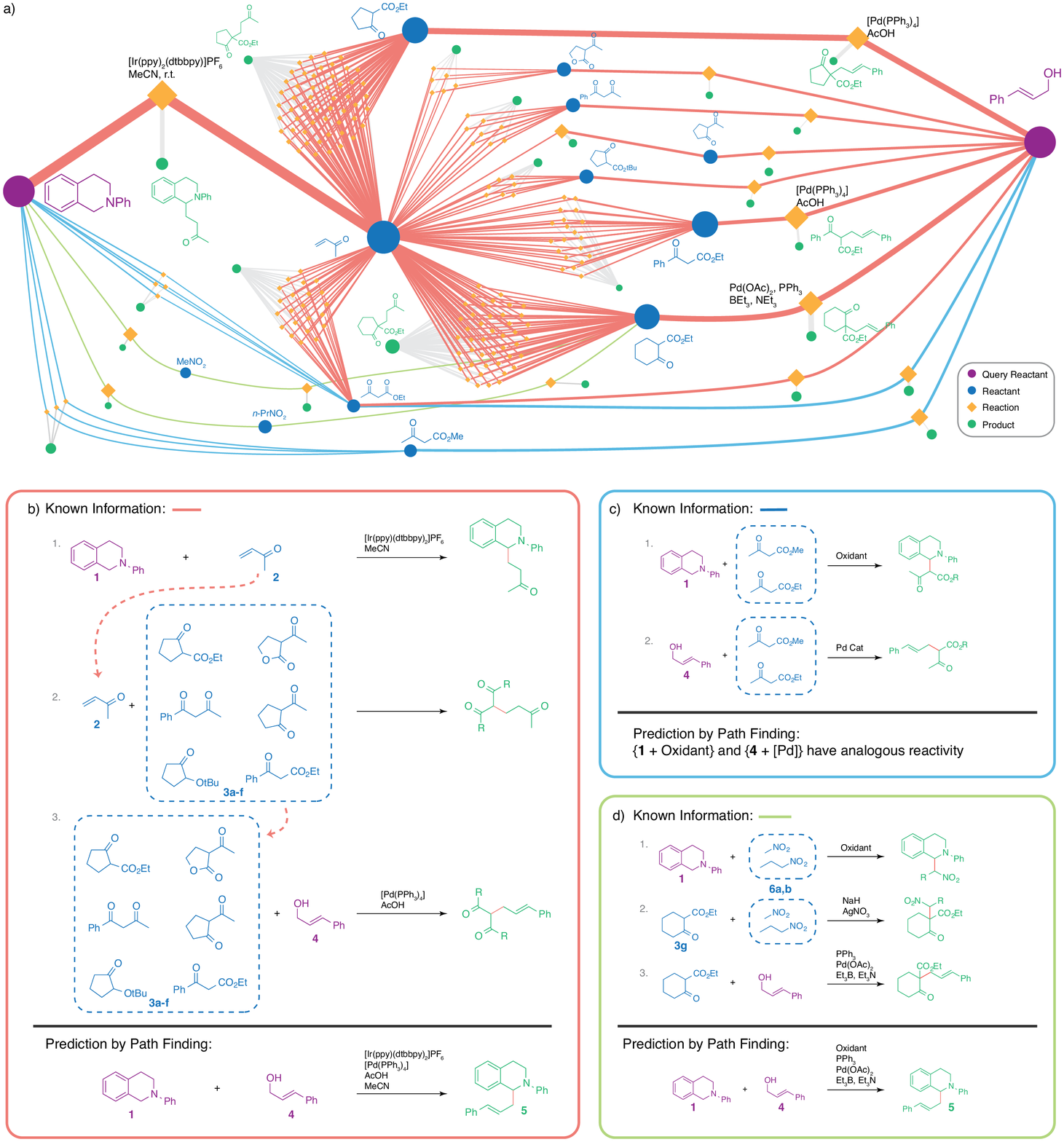}
\caption{a) Paths found between molecules 1 and 4. Many different paths are retrieved, which lead to three predictions (b,c,d): b) The red paths, which form the majority of paths between 1 and 4, lead to expected product 5. Amine 1 reacts with methyl vinyl ketone (MVK) 2 in the presence of a photoredox catalyst. MVK and allyl alcohol 4 both react with a variety of β-keto esters and therefore possess analogous reactivity. Since many paths are found, this prediction is supported by many different explanations.
c) The blue paths correspond to the prediction that 1 and 4 can also be analogous in their reactivity. 1 can be oxidised in the alpha-position to the nitrogen, which leads to an electrophilic imminium species, whereas 4 can act as an electrophile when activated with a Pd-catalyst. 
d) Even though the green paths give the same product as under b), the explanation is not plausible, because in step 1, the nitroalkanes 6a,b react as a nucleophile, while in step 2, they react under a different mechanism in an oxidative nucleophile coupling.}
\label{fig:xiao-scheme}
\end{center}
\end{figure}

\newpage
\mbox{}
\newpage

\subsection*{Interpreting Link Prediction as Chemical Reasoning}

Our approach has several interesting properties: 

1) Our model infers analogous or complementary reactivity of two molecules from their participation in known reactions. For example, our approach could infer that malonates and nitroalkanes show analogous behaviour towards aldehydes, even though they do not share the same substructure, or that phenyl bromide could act as a nucleophile in the presence of Magnesium, but as an electrophile in the presence of a Palladium catalyst, based only on reaction data. This is achieved without considering predefined functional groups, entered or extracted rules or expert knowledge. This is a big advantage, because defining how to extract reaction rules, and which neighbouring groups contribute to or impede reactivity is very difficult and entering large sets of reaction rules with functional group compatibility information by hand is cumbersome and labour intensive. Additionally, because our approach considers the complete molecules and not just isolated functional groups as in rules, information about functional group tolerance and regio-/chemoselectivity is implicitly encoded.

2) Essentially, our model performs reaction prediction by recombining the known reactions of the two query molecules. The path based search tells us which reactions we should choose for a chemically meaningful recombination. Paths in the knowledge graph can therefore be interpreted as a form of logical explanation. The retrieved paths can be directly inspected as an explanation of why a particular prediction is made. Unlike many machine-learning approaches, this makes our approach white-box, which is a desirable feature of artificial intelligence systems.\cite{ley2015organic}  However, we emphasise that the found paths are not a transitive law. Our model has to been seen as heuristic\cite{graulich2010heuristic} or as a form of inductive inference.\cite{klahr2002exploring,dunbar2005scientific,holyoak2012analogy,smith2012cognition} It can produce highly plausible predictions,  however the predictions are not foolproof. This is analogous to human generated hypotheses, where ideas that look good on paper might  turn out to not be experimentally feasible.\cite{graulich2010heuristic}

\begin{figure}[bt]
\begin{center}
\includegraphics[scale=0.6]{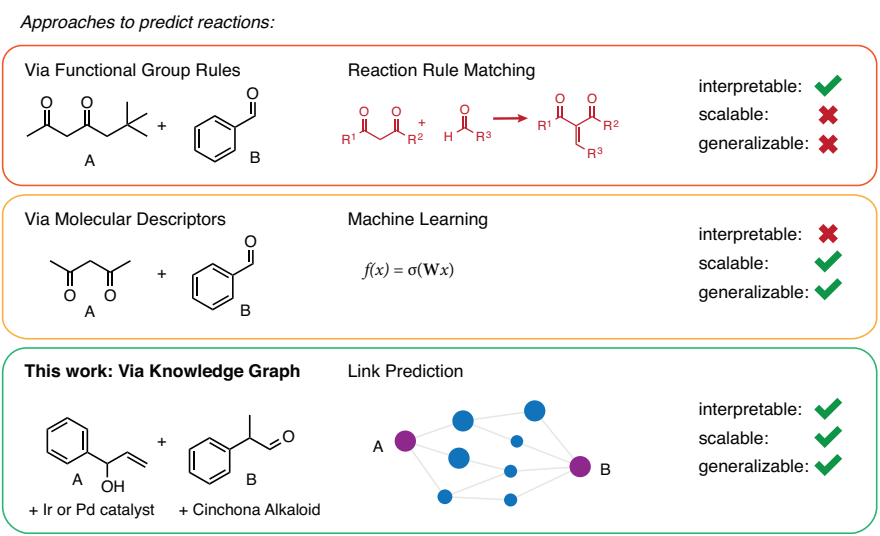}
\caption{An overview of methods available for high-throughput reaction prediction. }
\label{fig:mapping}
\end{center}
\end{figure}

In summary, our model compares favourably to rules based-expert systems or machine learning models for reaction prediction (see \ref{fig:mapping}).

\section*{Conclusion}
We have introduced a model that uses link prediction in a knowledge graph as a way of generating high quality hypotheses for reaction prediction and discovery. Through path-finding, our model detects how the reactivity of molecules are related, and if the reactivity is complementary or analogous. Using essentially the complete published knowledge of organic chemistry, we have quantitatively demonstrated that the model can successfully predict the products of binary reactions, and can also detect reactions that are unlikely to occur. Importantly, we have shown that the model can generalise beyond its knowledge, even to reaction types it does not know, a feature that current rule-based or machine learning systems do not possess. Empirical evaluation indicates that our approach can also be used to predict the catalysts and reagents for novel reactions, however, more quantitative research has to be conducted in this regard.

In this work, we have deliberately restricted the model to binary reactions and molecules with known reactions to introduce and validate the concept. Obviously, these limitations have to be overcome in the long term. Preliminary work in our laboratory indicates that these can be solved be an extension of the model, for example by augmenting the graph with abstract, hierarchical knowledge about molecules and reactions, similar to the hierarchical representations that human experts develop.\cite{holyoak2012analogy} In forthcoming work, we will furthermore examine how our ansatz can be combined with machine learning\cite{kayala2012reactionpredictor,zhang2005structure,marcou2015expert,ramakrishnan2015big} and low-cost quantum mechanics.\cite{finkelmann2016robust} 
Finally, we anticipate that our work will take up the pioneering work on computer-assisted reaction discovery by Ugi and Herges, and will be employed as an inspiration tool that provides ideas for unprecedented reactions.

\bibliography{mentor}

\section*{Acknowledgements}
This work was supported by Deutsche Forschungsgemeinschaft (SFB 858). We thank D. Evans (RELX Intellectual Properties SA) as well as J. Swienty-Busch and S. Radestock (Elsevier Information Systems GmbH) for the reaction dataset and ChemAxon for an academic software license.

\end{document}